\documentclass{article}

    \usepackage[preprint]{neurips_2021}

\usepackage[utf8]{inputenc} 
\usepackage[T1]{fontenc}    
\usepackage{hyperref}       
\usepackage{url}            
\usepackage{booktabs}       
\usepackage{amsfonts}       
\usepackage{nicefrac}       
\usepackage{microtype}      
\usepackage{xcolor}         

\newcommand{\name}{{\sc A-RTRS}}

\usepackage{mathrsfs} 
\usepackage{amsmath, amssymb}
\usepackage{graphicx}
\usepackage{subcaption}
\usepackage{enumerate}
\usepackage[short,nocomma]{optidef}
\usepackage{booktabs}

\title{Learning Model-Based Vehicle Relocation Decisions for Real-Time Ride-Sharing: \\
Hybridizing Learning and Optimization}

\author{%
  Enpeng Yuan \\
  School of Industrial and Systems Engineering \\
  Georgia Institute of Technology\\
  \texttt{eyuan8@gatech.edu} \\
   \And
   Pascal Van Hentenryck \\
   School of Industrial and Systems Engineering \\
   Georgia Institute of Technology \\
   \texttt{pvh@gatech.edu} \\
}

\begin{document}

\maketitle

\begin{abstract}
Large-scale ride-sharing systems combine real-time dispatching and routing optimization over a rolling time horizon with a model predictive control (MPC) component that relocates idle vehicles to anticipate the demand. The MPC optimization operates over a longer time horizon to compensate for the inherent myopic nature of the real-time dispatching. These longer time horizons are beneficial for the quality of relocation decisions but increase computational complexity. Consequently, the ride-sharing operators are often forced to use a relatively short time horizon. To address this computational challenge, this paper proposes a hybrid approach that combines machine learning and optimization. The machine-learning component learns the optimal solution to the MPC on the aggregated level to overcome the sparsity and high-dimensionality of the solution. The optimization component transforms the machine-learning prediction back to the original granularity through a tractable transportation model. As a consequence, the original NP-hard MPC problem is reduced to a \emph{polynomial time} prediction and optimization, which allows the ride-sharing operators to consider a longer time horizon.
Experimental results show that the hybrid approach achieves significantly better service quality than the MPC optimization in terms of average rider waiting time, due to its ability to model a longer horizon.


\end{abstract}

\section{Introduction}

Rapid growth of ride-hailing market in recent years has greatly
transformed urban mobility, offering on-demand mobility services via
mobile application. While major ride-hailing platforms such as Uber
and Didi leverage centralized dispatching algorithms to find good
matching between drivers and riders, operational challenges still
persist due to imbalance between demand and supply. Consider morning
rush hours as an example: most trips originate from residential areas
to business districts where a large number of vehicles accumulate and remain idle. Relocating these vehicles back to the demand area is thus
crucial to maintaining quality of service and income for the drivers.

There has been extensive studies on vehicle relocation problem in
real-time. Previous methodologies fit broadly into two categories:
optimization-based approach and learning-based
approach. Optimization-based approach involves solving a mathematical
program using expected demand and supply information over a future horizon to derive
relocation decisions. Learning-based approach (predominantly
reinforcement learning) trains a state-based decision policy from
interacting with the environment and observing the rewards. While both
approach have demonstrated promising performance in simulation and (in
some cases) real-world deployment \citep{Didi}, they have obvious
drawbacks - the optimization needs to be solved in real-time and often
trades off fidelity (hence quality of solutions) for computational
efficiency. Reinforcement-learning approaches require a tremendous
amount of data to explore high-dimensional state-action spaces and
often simplify the problem to ensure efficient training. While a
complex real-world problem like vehicle relocation may never admit a
perfect solution, there are certainly possibilities for improvement.

This paper presents a step to overcoming these computational
challenges. It proposes to replace an existing relocation optimization by a
machine-learning model that predicts the optimal solutions. This
makes it possible for the real-time framework to consider the optimization at higher fidelity (since the optimizations can be solved and
learned offline) and therefore improve the overall quality of service.
This learning step however comes with several challenges. First, the
decision space is usually high-dimensional (e.g., number of vehicles
to relocate between each pickup/dropoff location) and sparse, since
relocations typically occur only between a few low-demand and
high-demand regions. Capturing such patterns is difficult even with
large amount of data. Second, the predicted solutions may not be
feasible, as most predictive algorithms cannot enforce physical
constraints that the solutions need to satisfy.

To solve these technical difficulties, the paper proposes an
aggregation-disaggregation procedure, which learns the
relocation decisions at an aggregated level to overcome the high
dimensionality and sparsity of the data, and convert them back to
feasible solutions at the original granularity by a
transportation optimization that can be solved in polynomial
time. \emph{As a consequence, in real time, the original NP-Hard
  optimization problem is reduced to a polynomial-time problem of
  prediction and optimization}. 


The proposed learning/optimization framework is evaluated on
New York Taxi data set and achieves a $27\%$ reduction in average waiting time
compared to the original relocation model due to its higher fidelity. The results suggest that {\em a
  hybrid approach, that combines a high-fidelity optimization model for
  real-time dispatching and routing decisions and a machine-learning
  model for tactical relocation decisions, may provide an appealing avenue for
  certain classes of real-time optimizations.}

The paper is organized as follows. Section \ref{Sec:statement} defines
the relocation problem, Section \ref{Sec:prior_works} summarizes the
existing literature, Section \ref{Sec:reloc_model} reviews the
relocation model, Section \ref{Sec:methodology} presents the
aggregation-disaggregation learning framework, and Section
\ref{Sec:Expriments} reports the experimental results.

\section{Problem Statement}
\label{Sec:statement}

\begin{figure}[t]
    \centering
    \includegraphics[width = \linewidth]{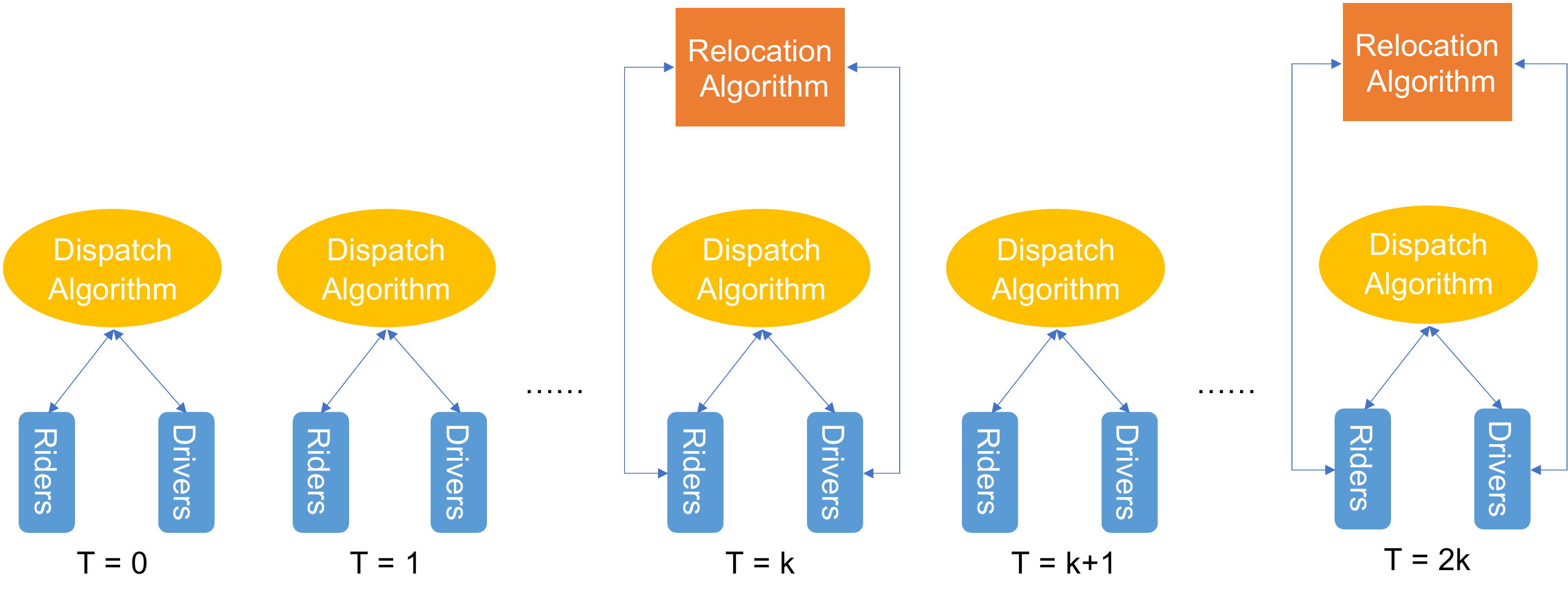}
    \caption{The real-time ride-sharing operations.}
    \label{fig:big_pic}    
\end{figure}

This paper considers a ride-sharing platform that uses a fixed fleet
of autonomous vehicles or a pool of drivers who follow relocation
instructions exactly. The platform runs a dispatching algorithm at high frequency (every 5 - 30 seconds) to match drivers with riders and a relocation algorithm at lower frequency (every 5 - 10 minutes) to relocate idle vehicles back to the demand areas (See Figure \ref{fig:big_pic} for an illustration of the overall framework). Due to its forward-looking nature, the relocation optimization makes decisions based on expected demand and supply in a future time window. Longer time windows improve the quality of decision but make it harder to solve the problem in real-time. Therefore this paper learns the optimal solution of a relocation optimization with a sufficiently long horizon, and demonstrates that the learned policy achieves similar performance as the optimization model in terms of average rider waiting time and relocation costs.

\section{Prior Work}
\label{Sec:prior_works}

Prior results on real-time idle vehicle relocation fit broadly into
two frameworks: model predictive control (MPC) (\citep{Miao, Zhang,
  Iglesias2017, Huang, ijcai2020}) and reinforcement learning (RL)
(\citep{Aug, MOVI, SAMoD, MARL, CoRide, Didi, CP}). MPC is an online
control procedure that repeatedly solves an optimization problem over
a moving time window to find the best control action. System dynamics
are explicitly modeled as mathematical constraints. Due to
computational complexity, all the MPC models in the literature work at
discrete spatial-temporal scale (dispatch area partitioned into zones,
time into epochs) and cover a relatively short time horizon.

Reinforcement learning, on the contrary, does not explicitly model
system dynamics and trains a decision policy offline by approximating
the state/state-action value function. It can be divided
into two streams: single-agent RL \citep{Aug, MOVI, SAMoD, CP}, and
multi-agent RL \citep{MARL, CoRide, Didi}. Single-agent RL focuses on
maximizing reward of an individual agent, and multi-agent RL maximizes
collective rewards of all the agents. The main challenge of this approach
is to efficiently learn the state-action value function, which is
high-dimensional (often infinite-dimensional) due to the complex and
fast-changing demand-supply dynamics that arises in real-time
settings. Since RL relies solely on interacting with the environment
to approximate the value function, a tremendous amount of samples need
to be generated to fully explore the state-action space. Consequently,
many works simplify the state-action space by using the same policy
for agents within the same region \citep{Aug, MARL}, or restrict
relocations to only neighboring regions \citep{MOVI, SAMoD, CoRide,
  Didi}.

Our approach tries to combine the strength of both worlds - it models the
system dynamics explicitly through a sophisticated MPC model, and
approximates the optimal solutions of the MPC by machine learning to
overcome the real-time computational challenge. As far as the authors
know, the only work that has taken a similar approach is \citet{Lei},
where the authors propose to learn the decisions of a relocation model
and show that the learned policy performs close to the original
model. However, their model considers only one epoch (10
mins) and does not track how demand and supply interact over an
extended period of time. This paper focuses on a much more sophisticated MPC
with multiple epochs and demonstrates that the learned policy
achieves \textbf{superior} performance than the original model due to its ability to model a longer horizon within the computational limits.


\section{The Relocation Model}
\label{Sec:reloc_model}

The underlying relocation model is an MPC model that operates over a moving time horizon. Specifically, time is discretized into epochs of equal length and, during each epoch, the
MPC performs three tasks: (1) it predicts the demand and supply for the next $T$
epochs; (2) it optimizes decisions over these epochs; and (3) it
implements the decisions of the first epoch. In addition, space is partitioned into zones (not necessarily of equal size or shape) and relocation decisions are made on the zone-to-zone level for each epoch. It is assumed that vehicles only pick up demand in the same zone. Once vehicles start to deliver riders
or relocate, they must finish the trip before taking another
assignment. These assumptions help the MPC \emph{approximates} how the
underlying routing/dispatch algorithm works in reality, but the
dispatch algorithm does not have to obey these constraints. The only
interaction between the routing/dispatch algorithm and the MPC is
due to the relocation decisions. To account for waiting time, riders
can only be picked up in $s$ epochs since they arrive in the system:
they are assumed to drop out if waiting more than $s$ epochs.

The model formulation is presented in Figure \ref{fig:MPC}. $Z$ denotes
the set of zones and $\mathcal{T}=\{1,2,...,T\}$ set of time epochs in
the time window. In the rest of the paper, we use $i$ and $j$ to
denote zone index, and $t$ and $\rho$ to denote time index. Decision
variables are as follows: $x^p_{ijt\rho}$ denotes number of vehicles
starting to serve riders from $i$ to $j$ in $\rho$ who arrived in
$t$. $x_{ijt}^r$ denotes the number of vehicles starting to relocate from
$i$ to $j$ during $t$ ($j\neq i$), $z_{it}$ denotes the number of vehicles
staying in $i$ from $t-1$ to $t$, and $l_{it}$ indicates whether there
is demand unserved in zone $i$ at the end of epoch $t$.

The model inputs are the following. $D_{ijt}$ denotes number of
vehicles needed to serve expected riders between $i$ and $j$ arriving
at $t$. $V_{it}$ denotes number of expected idle vehicles in zone $i$
in $t$: those vehicles are busy now but will become available in $i$ in $t$. $\lambda_{ij}$ denotes number of epochs to travel from $i$
to $j$. Furthermore, $\phi(t) = \{\rho\in\mathcal{T}: t\leq \rho \leq
t+s-1\}$ denotes set of valid pick-up epochs for riders arriving in
epoch $t$.

Constraint (\ref{constr_demand1}) makes sure that vehicles do not
serve more demand than there is in the allowed waiting time
limit. Constraints (\ref{constr_init_idle}) and (\ref{constr_balance})
are flow balance constraints for each zone and epoch. Big-M
constraints (\ref{no_relocate_constraint1}) and
(\ref{no_relocate_constraint2}) prevent vehicles from relocating
unless all demand in the current zone is served. The objective
maximizes a weighted sum of total riders served and minimizes the
total relocation cost. $W_{ij}$ is the average number of riders from
$i$ to $j$ a vehicle carries: a vehicle can pick up multiple
ride-sharing requests or a single request with multiple passengers,
which can be estimated from historical data and taken as a
parameter. $q^p(t,\rho)$ is the weight of a rider served at $\rho$ who
arrives at $t$, and
$q^r_{ij}(t)$ is the relocation cost between $i$ and $j$ in $t$. $q^p(t,\rho)$ and $q^r_{ij}(t)$ are decreasing in $t$ 
and $\rho$ since uncertainty about the future grows over time.

The model is a Mixed-Integer Linear Program (MILP), which is
\emph{NP-Hard} and challenging to solve at high-fidelity when the
number of zones $|Z|$, or the time horizon $|\mathcal{T}|$, are large.

\begin{figure}[!t]
\small
\begin{maxi!}[2]<b>
    {}{\underset{i,j}\sum \underset{t,\rho} \sum q^p(t,\rho) W_{ij}x^p_{ijt\rho} - \underset{i,j} \sum \underset{t} \sum q^r_{ij}(t) x^r_{ijt}}
    {}{} \notag
    \addConstraint{\sum_{\rho\in\phi(t)} x^p_{ijt\rho}}{ \leq D_{ijt},} {\forall i,j,t}\label{constr_demand1}
    \addConstraint{z_{i1}}{=0,}{\forall i}  \label{constr_init_idle}
    \addConstraint{\sum_{j, t_0} x^p_{ijt_0 t} + \sum_{j} x^r_{ijt} + z_{i (t+1)} }{= \sum_{j, t_0} x^p_{jit_0 (t - \lambda_{ji})} + \sum_{j} x^r_{ji(t - \lambda_{ji})} + z_{i t} + V_{i t}, \quad }{\forall i,t} \label{constr_balance} 
    \addConstraint{\sum_{j} x^r_{ijt}}{\leq M l_{it},}{\forall i,t} \label{no_relocate_constraint1}
    \addConstraint{\sum_j \sum_{t_0: \, t\in\phi(t_0)} \left(D_{ijt} - \sum_{\rho=t_0}^{t} x^p_{ijt_0\rho} \right)  }{\leq M(1 - l_{it}),}{ \forall i,t} \label{no_relocate_constraint2}
    \addConstraint{x^p_{ijt\rho}, x^r_{ijt}, z_{it}}{\in \mathbb{Z_+},}{\forall i,j,t,\rho} \label{constr_var_type1}
    \addConstraint{l_{it}}{\in\{0,1\},}{\forall i,t} \label{constr_var_type2}
\end{maxi!}
\caption{The MPC relocation model}
\label{fig:MPC}
\end{figure}

\section{The Learning Methodology}
\label{Sec:methodology}

This section presents the overall framework for learning relocation
decisions. It consists of four steps:
\begin{enumerate}
  \item {\em Aggregation:} the decisions of the MPC model are first aggregated to the
    zone-level for efficient learning;

  \item {\em Learning:} the aggregated decisions are learned;

  \item {\em Feasibility Restoration:} the learned aggregated
    decisions are post-processed to restore feasibility;

  \item {\em Disaggregation:} the post-processed aggregated decisions
    are disaggregated to the original zone-to-zone level through a
    transporation optimization.
\end{enumerate}

\noindent
Section \ref{Sec:aggreg} specifies the aggregation procedure, Section
\ref{Sec:learning_method} presents the learning techniques, and
Sections \ref{Sec:rounding} and \ref{Sec:transport} present the method
to restore feasibility and the disaggregation model.



\subsection{The Aggregation Component}
\label{Sec:aggreg}

The overarching goal of the framework is to learn the decisions of the relocation MPC model in the first
epoch, i.e., $[x^r_{ij1}]_{i,j\in Z}$, since these are the decisions
implemented by the platform after each MPC run. In reality, the
decision vector $x^r_{ij1}$ is often sparse, with relocation occurring
only between a few low-demand zones and high-demand zones. To
reduce the sparsity and the dimension of the output,
$x^r_{ij1}$ are first aggregated to the zone level. More precisely, two metrics will be learned for each zone $i$ - the
number of vehicles relocating from $i$ to other zones, i.e., $y^o_i:=
\sum_{j\in Z, j\neq i} x^r_{ij1}$, and the number of vehicles
relocating to $i$ from other zones, i.e., $y^d_i := \sum_{j\in Z,
  j\neq i} x^r_{ji1}$. Note that these two metrics can be both
non-zero at the same time - an idle vehicle might be relocated from $i$
to another zone to serve a request in the near future,
and another vehicle could come to $i$ to serve a request that will arise
much later. Therefore the output dimension is reduced from $|Z|^2$ to
$2|Z|$.

\subsection{The Learning Component}
\label{Sec:learning_method}

The learning algorithm takes essentially the same input as the MPC:
the predicted supply $\mathbf{V} =[V_{it}]_{i\in Z,
  t\in\mathcal{T}}$ and the predicted demand $\mathbf{D} =
[D_{ijt}]_{i,j\in Z, t\in\mathcal{T}}$ of idle vehicles. The aggregated relocation decisions
$\mathbf{y} = [y^d_i, y^o_i]_{i\in Z}$ are predicted from
$\mathbf{V}$ and $\mathbf{D}$.

The trip demand in the real world often exhibits large variations. For
example, business areas sees high volume of demand during weekdays and
few requests on the weekend. Special events such as concerts or
extreme weather could also lead to atypical demand patterns. This
often causes the demand distribution and the corresponding
relocation decisions to follow long-tail distribution. It is necessary to resample/reweigh the data to create a balanced data set. However, common sampling and weighting techniques are designed
for univariate labels whereas $\mathbf{y}$ is a
vector. One possibility is to sample the data and train a different
model for each element in the label. However, this increases the
training complexity and space requirements for deployment, especially
when the underlying model has a large number of parameters (e.g., deep neural
network). Therefore, we propose a mean-sampling and weighting heuristic
to balance the data for all elements. First, the data is sampled based
on the label mean: if the mean of a label falls within a predefined tail
range, the observation is over-sampled. This does not guarantee,
however, that the distributions of all individual elements are
balanced, since each element in the label may follow a different
distribution. The second step identifies elements
whose distributions are still unbalanced after mean-sampling, and
increases these elements' weights in the objective function when they fall into
the tail range. We experimentally demonstrate in Section
\ref{Sec:Expriments} that this procedure learns each element
accurately.





\subsection{Feasibility Restoration}
\label{Sec:rounding}
The feasibilty restoration turns the predictions $\mathbf{\hat{y}} =
[\hat{y}^o_i, \hat{y}^d_i]_{i\in Z}$ into feasible relocation
decisions that are integral and obey the (hard) flow balance
constraints. This is performed in three steps. First, $\hat{y}^o_i$
and $\hat{y}^d_i$ are rounded to their nearest non-negative
integers. Second, to make sure that there are not more relocations
than idle vehicles, the restoration sets $\hat{y}^o_i =
\min\{\hat{y}^o_i, V_{i1}\}, \; \forall i\in Z$, where $V_{i1}$ is
number of idle vehicles in zone $i$ in the first epoch. Finally,
$\hat{y}^o_i$ and $\hat{y}^d_i$ need to satisfy flow balance
constraint, e.g., $\sum_{i\in Z} \hat{y}^o_i = \sum_{i\in Z}
\hat{y}^d_i$. This is ensured by setting the two terms to be the
minimum of the two, by randomly decreasing some non-zero elements
of the larger term.

\begin{figure}[t]
\begin{align*}
    \min \quad & \sum_{i,j\in Z} c_{ij} z_{ij} \notag \\
    \text{s.t.} \quad & \sum_{j\in Z} z_{ij} = \hat{y}^o_i, \quad & \forall i\in Z\\ 
    & \sum_{j\in Z} z_{ji} = \hat{y}^d_i, \quad & \forall i\in Z\\
    & z_{ij} \in \mathbb{Z^+}, \quad & \forall i,j\in Z
\end{align*}
\caption{The transportation problem}
\label{fig:ASSIGN}
\end{figure}

\subsection{Disaggregation Through Optimization}
\label{Sec:transport}

The previous steps produce a feasible relocation plan
$[\hat{y}^d_i,\hat{y}^o_i]_{i\in Z}$ on the zone level. This last step reconstructs it to the zone-to-zone level via a transportation optimization. The model formulation is given in Figure
\ref{fig:ASSIGN}. Variable $z_{ij}$ denotes the number of vehicles to
relocate from $i$ to $j$, and constant $c_{ij}$ represents the
corresponding relocation cost. The model minimizes the total
relocation cost to consolidate the relocation plan. Its solution
$z_{ij}$ will be implemented by the dispatching platform in the same
way as $x^r_{ij1}$ was for the MPC. This optimization is a form of
balanced transportation optimization problem ($\sum_i \hat{y}^d_i =
\sum_i \hat{y}^o_i$): it is totally unimodular and can be solved in
\emph{polynomial time} by solving its linear programming
relaxation. In addition, the problem is always feasible, and does not
depend on the length of MPC time horizon. Therefore, the original NP-Hard MPC has
been reduced to a polynomial time algorithm that transforms the predictions
into a near-optimal solution to the original problem.

\section{Experimental Results}
\label{Sec:Expriments}

The proposed learning framework is evaluated on Yellow Taxi Data in
Manhattan, New York City \citep{nycdata}. Several representative
instances between 2015 and 2016 are selected for training, and the
learned policy is tested on data during February, 2017. Section
\ref{Sec:simu} reviews the simulation environment, Section
\ref{Sec:train} presents the learning results, and Section
\ref{Sec:results} evaluates the learned policy's performance on the
test cases.

\subsection{Simulation Environment}
\label{Sec:simu}
\begin{figure}
    \centering
    \includegraphics[width = 0.95\linewidth]{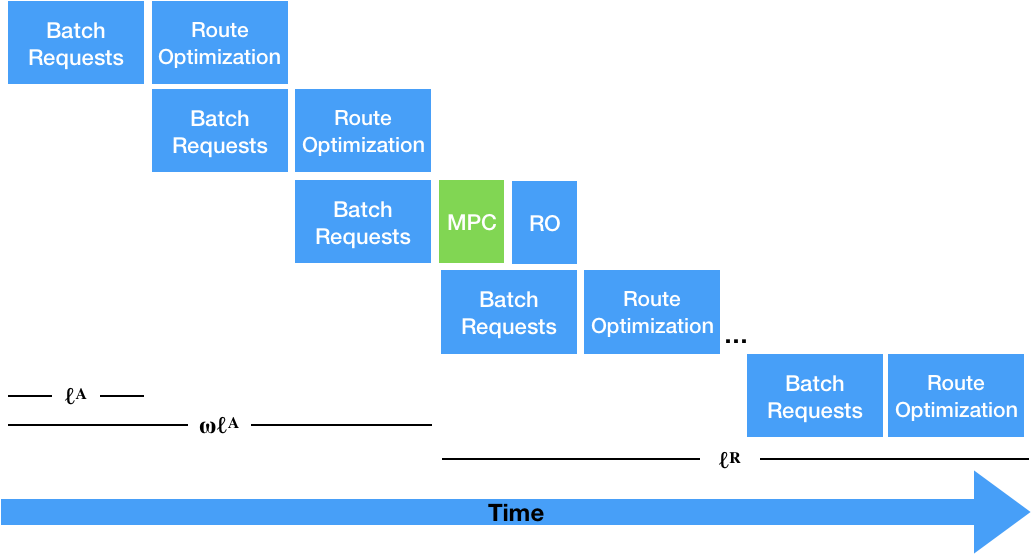}
    \caption{The \name{} framework}
    \label{fig:mpc_illustrated}
\end{figure}

The end-to-end ride-sharing simulator \name{} (illustrated in Figure \ref{fig:mpc_illustrated}) is used to generate data and evaluate relocation policies \citep{ijcai2020}. The Manhattan area is partitioned into a grid of cells of $200$ squared
meter and each cell represents a pickup/dropoff location. Travel time
between the cells are queried from \citet{OpenStreetMap}. The fleet is
fixed to be $2000$ vehicles with capacity $4$, distributed randomly
among the cells at the beginning of the simulation. The simulator uses
a ride-sharing routing and dispatching algorithm that batches requests
into a time window and optimizes every 30 seconds
\citep{riley2019}. Its dispatching objective is to
minimize a weighted sum of passenger waiting times and penalties on
unserved requests. Each time a
request is not served, its penalty is increased in the next batch window to ensure that the request will be scheduled with higher priority. The algorithm is
solved by column generation: it iterates between solving a restricted
master problem (RMP), which assigns a route to each vehicle, and a
pricing subproblem, which generates feasible routes for the vehicles. The relocation MPC (Figure \ref{fig:MPC}) is executed every
$5$ minutes. After the MPC decides zone-to-zone level relocations, a
vehicle assignment optimization determines which individual vehicles to relocate by minimizing total traveling distances. Of the three models, the dispatching model is the most computationally
intensive since it operates on the individual (driver and rider) level
as opposed to the zone level. Since all three models must be executed in the 30 seconds batch window,
the platform allocates $20$ seconds to the dispatching optimization, $5$ seconds to the relocation optimization, and $5$ seconds to the
vehicle assignment. All the
models are solved using Gurobi 9.0 \citep{gurobi}.

\subsection{Learning Results}
\label{Sec:train}

\subsubsection{Training Data}

24 taxi instances between 2015 and 2016 were selected as
representative instances for learning. Each instance runs from 7 am to
9 am, with the total number of riders ranging from 19,276 to 59,820,
representing a wide variety of demand scenarios in Manhattan. The
instances are then perturbed by randomly adding/deleting certain
percentage of requests to generate more training instances, where the
percentages are sampled from a normal distribution $N(0, 2.5)$. The
instances are run by the simulator and the MPC results are extracted
as training inputs. The MPC setup is as follows: It partitions the
area into $73$ zones and time into $5$-minute epochs. The MPC time
window contains $6$ epochs. Riders can be served in $3$
epochs following their arrival. Demand predictions for each O-D pair
and epoch is generated by adding white noise to the true demand. The
white noise is normally distributed with zero mean and a standard
deviation equal to $2.5\%$ of the true demand. The number of idle
vehicles in each epoch is estimated by the simulator based on current
route of each vehicle and the travel times. The ride-share ratio is 
$W_{ij}=1.5$ for all $i,j\in Z$. Service weight and
relocation penalty are $q^p(t,\rho)=0.5^{t}0.75^{\rho-t}$ and
$q^r_{ij}(t)=0.001*0.5^{t}\eta_{ij}$ where $\eta_{ij}$ is travel time
between zone $i$ and zone $j$ in seconds.

The MPC at such spatio-temporal fidelity is difficult to solve. With 6
cores of 2.1 GHz Intel Skylake Xeon CPU, $24.7\%$ instances could not
find a feasible solution within the 5-second solver time. This was
the key motivation to explore a machine-learning approach. 

\subsubsection{Training}

\begin{figure}[!t]
    \minipage{0.5\textwidth}
    \includegraphics[width=\linewidth]{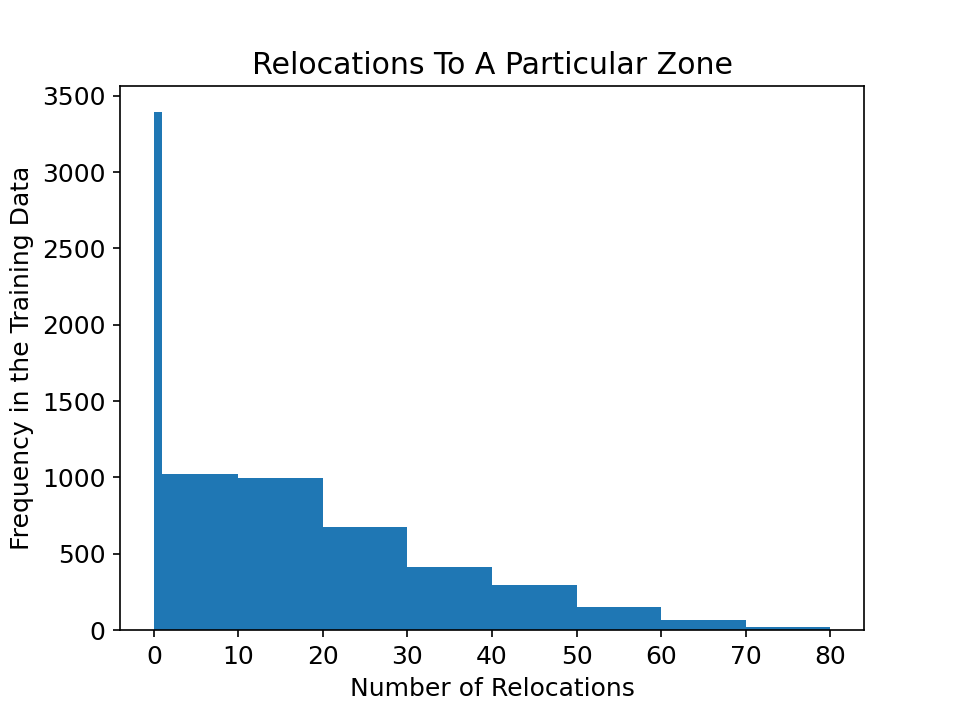}
    \caption{The distribution of relocations.}
    \label{fig:long_tail}
    \endminipage\hfill
    \minipage{0.5\textwidth}
    \includegraphics[width=\linewidth]{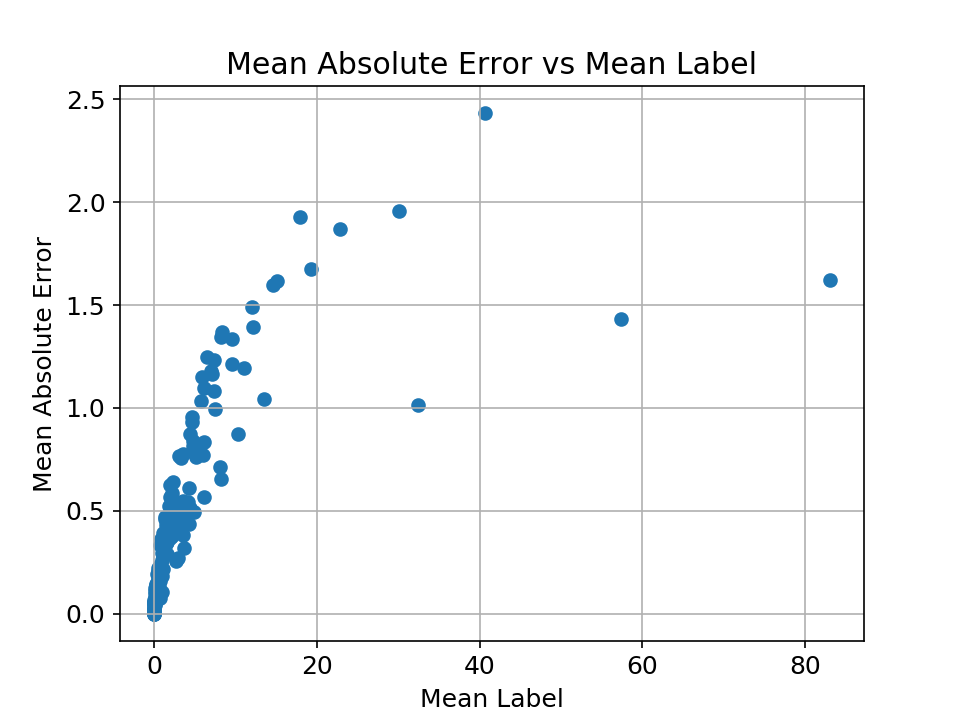}
    \caption{MAE vs mean label.}
    \label{fig:mae}
    \endminipage\hfill
\end{figure}

The learning model is an artificial neural network with mean
squared error (MSE) loss and $l_1$-regularization. As mentioned
earlier, the data input and output often follow long-tail
distribution, as highlighted in Figure \ref{fig:long_tail} which shows distribution of the number of vehicles relocating to a particular
zone. The mean-sampling and weighting procedure is performed in the
following way: if the mean of a training label is above a certain
threshold $\kappa$, the training example is duplicated $\mu$ times in
the training set where $(\kappa, \mu) = (4.0, 3)$ in the
experiment. After the mean-sampling, some
elements of the label that are zero in most training examples still exhibit a
long-tail pattern. The weights of these sparse elements in
the MSE loss are multiplied by a factor of $5$ on instances where they take non-zero
values. This makes the training balanced for all elements.

The artificial neural network contains two fully-connected layers with hyperbolic tangent activation function and 1024 hidden units
each. It is trained by the Adam optimizer in Pytorch with batch size
$32$ and learning rate $10^{-3}$ \citep{Adam, PyTorch}. The model is
trained on 10,000 data points and validated on 2000 data points. The
predictions are rounded to feasible solutions by
the procedure described in \ref{Sec:rounding}. Element-wise mean
squared error on the validation set is reported in Table
\ref{tab:val_error}, and the mean absolute error of each label element as
well as its mean on the validation set is given in Figure
\ref{fig:mae}. The error for each element is reasonable, indicating that the model
successfully learns the relocation pattern for each zone.

\begin{table}[b]
    \caption{The validation error before and after rounding.}
    \centering
    \vspace{6pt}
    \begin{tabular}{c r r }
        \toprule
        Model & MSE (Pre-Rounding) & MSE (Post-Rounding)\\
        \midrule
        DNN & 2.01 & 2.05 \\
        \bottomrule
    \end{tabular}
    \label{tab:val_error}
\end{table}

\subsection{Experimental Results}
\label{Sec:results}
\begin{figure}[!t]
    \minipage{0.33\textwidth}
    \includegraphics[width=\linewidth]{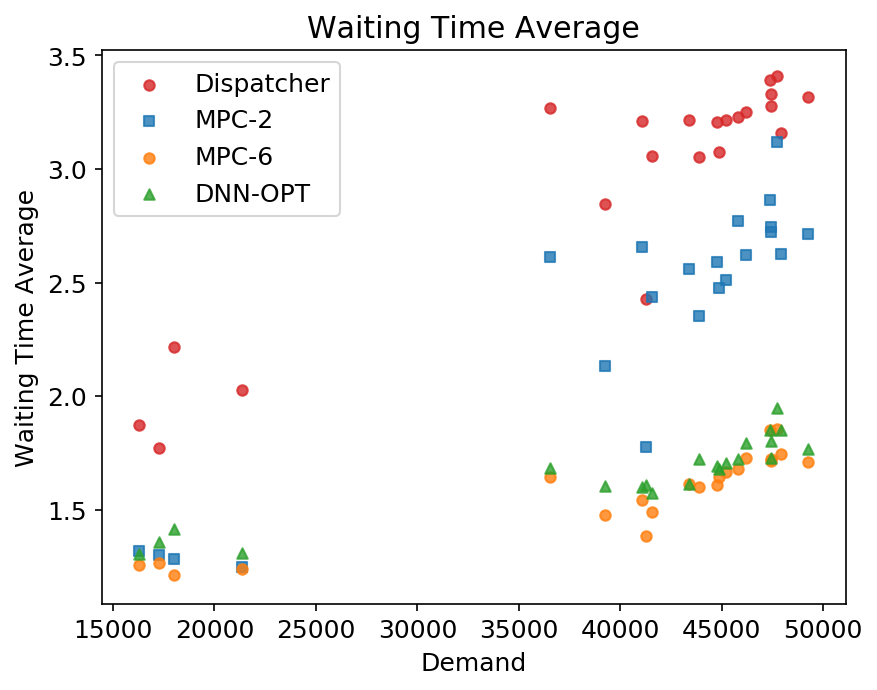}
    \subcaption{Waiting time averages.}
    \label{fig1a}
    \endminipage\hfill
    \minipage{0.33\textwidth}
    \includegraphics[width=\linewidth]{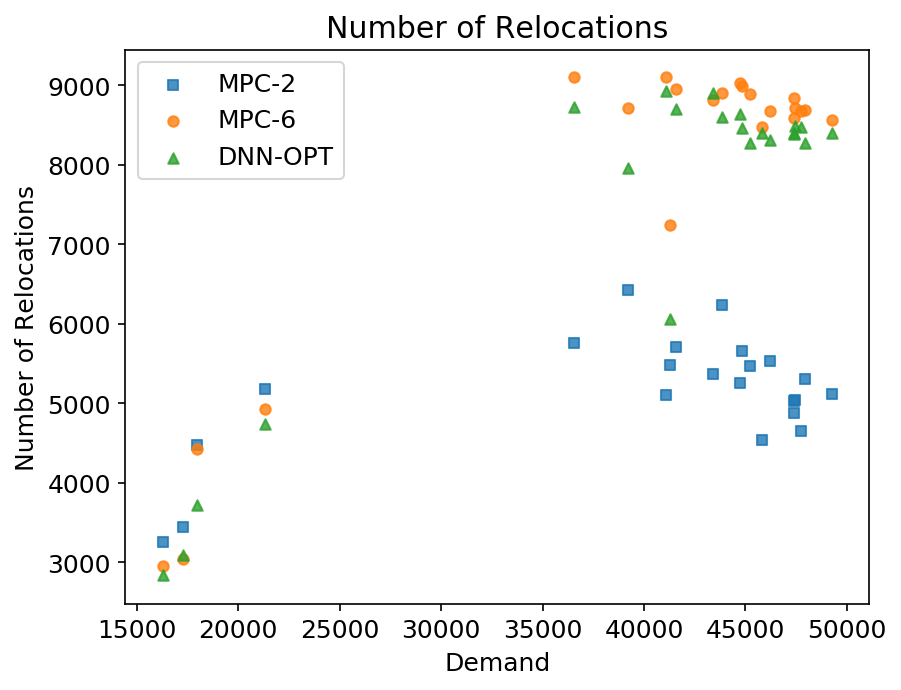}
    \subcaption{Number of relocations.}
    \label{fig1b}
    \endminipage\hfill
    \minipage{0.33\textwidth}
    \includegraphics[width=\linewidth]{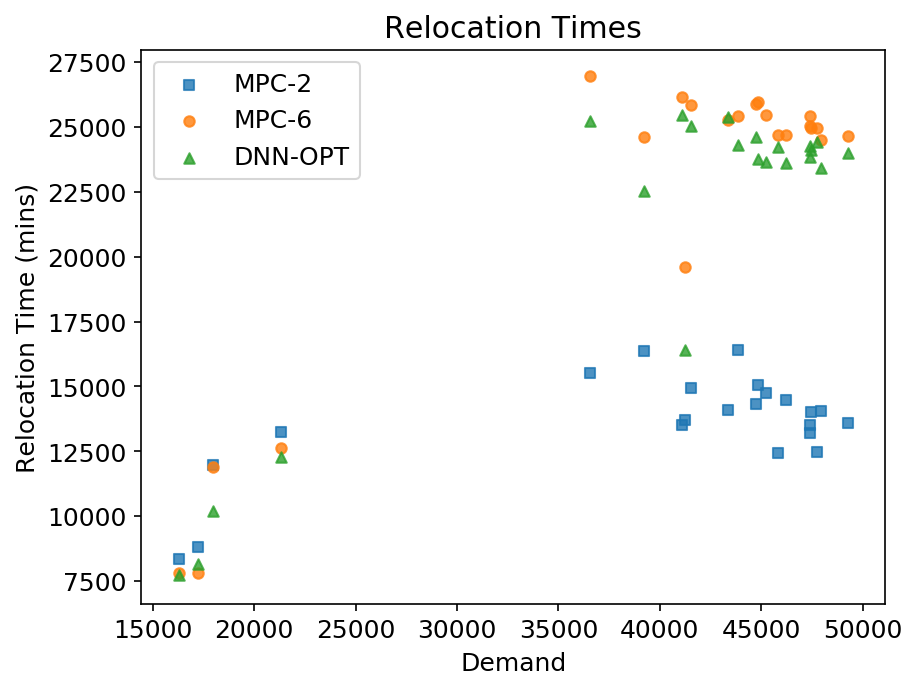}
    \subcaption{Relocation times.}
    \label{fig1c}
    \endminipage\hfill
    \caption{Evaluation results of DNN-OPT and its comparison with pure optimization approaches.}
    \label{fig1}
\end{figure}

The trained policy is evaluated on Yellow taxi data in February,
2017. To match with the training data distribution, days with
less than $16,000$ riders between 7am and 9 am are filtered out, which leaves $22$
days. The trained policy (DNN-OPT) is compared with the original
MPC model with $6$ epochs in the planning horizon (MPC-6), an MPC
model with 2 epochs that represents what can be solved within the
computational limit (MPC-2), and a baseline that does not perform relocations (Dispatcher). All the models use the same dispatching and routing algorithm introduced in \ref{Sec:simu}. The rider waiting time averages,
the number of relocations, and the total relocation time are reported
in Figure \ref{fig1}. In all instances, the waiting times obtained by
DNN-OPT is close to that of MPC-6, and significantly better than
Dispatcher and MPC-2. The reduction in average waiting time for MPC-2, MPC-6, and DNN-OPT compared to Dispatcher are $18.7\%, 47.8\%, 45.6\%$ on instances with more than $25,000$ riders. This shows that DNN-OPT achieves $27\%$ further reduction in waiting time than the MPC model within the computational limit. In terms of relocations, the total number of relocations and the relocation times of DNN-OPT are also close to that of MPC-6, indicating that DNN-OPT can effectively approximate the decisions of the original model. More relocations are performed under
MPC-6 and DNN-OPT than MPC-2 since longer time horizon unveils more
information about the future and more opportunity for relocations. The
solver times of the transportation optimization are reported in Table
\ref{tab:trans_time}, indicating that it can be solved
efficiently. The prediction time is also within fraction of a
second. {\em Overall, these promising results demonstrate that the proposed framework is capable of looking at longer time horizons for relocation, which leads to significant improvements in service quality.}

\begin{table}
\caption{The transportation model run times.}
\vspace{6pt}
\centering
    \begin{tabular}{c c c }
        \toprule
        & Mean & Max\\
        \midrule
        Solver Time (s) & 0.02 & 0.05 \\
        \bottomrule
    \end{tabular}
\label{tab:trans_time}
\end{table}

\section{Conclusion}

Large-scale ride-sharing systems often combine routing/dispatching with idle vehicle relocation. The relocation optimization is based on expected demand and supply in a future time horizon. Longer time horizons improve the quality of decision but make it harder to solve the model in real-time. This paper proposes a hybrid approach combining machine learning and optimization to tackle this computational challenge. The learning component learns the optimal solution of a sophisticated relocation optimization on the aggregated (zone) level, and the optimization component transforms the prediction back to feasible solution on the zone-to-zone level via a polynomial-time transportation problem. As a consequence, the original NP-hard optimization is reduced to a polynomial time prediction and optimization, which allows the use of longer time horizons. Simulation experiments on New York Taxi data demonstrate that the learned policy achieves significantly better service quality compared to the original optimization due to its longer horizon. In particular, the proposed approach further reduces average rider waiting time by 27\%. Future work is focusing on learning other important ride-sharing decisions such as dynamic pricing.

\bibliographystyle{abbrvnat}
\bibliography{main}

\newpage

\end{document}